\documentclass[11pt, a4paper, twocolumn]{article}
\usepackage[utf8]{inputenc}
\usepackage{multicol}
\usepackage{fancyhdr}
\usepackage{setspace}
\usepackage{indentfirst}
\usepackage{amsmath}
\usepackage{graphicx}
\usepackage{caption}
\usepackage{pxfonts}
\usepackage[subrefformat=parens,labelformat=parens]{subcaption}
\usepackage[a4paper]{geometry}
\usepackage{hyperref}
\usepackage[backend=bibtex,style=numeric-comp,sorting=none,firstinits=true,maxbibnames=99]{biblatex}
\DeclareNameAlias{author}{last-first}
\bibliography{reference}

\usepackage{sectsty} 
\sectionfont{\fontsize{12}{15}\selectfont}
\subsectionfont{\fontsize{12}{15}\selectfont}

\usepackage{lipsum}  

\makeatletter
\renewcommand{\maketitle}{\bgroup\setlength{\parindent}{0pt}
\begin{flushleft}
  \onehalfspacing
  \fontsize{20}{23}\selectfont
  \textbf{\@title} \\
  \hfill \break
  \fontsize{12}{15}\selectfont
  \@author
\end{flushleft}\egroup
}
\makeatother

\title{Fast Neural Network based Solving of Partial Differential Equations}
\author{
        \textbf{Jaroslaw Rzepecki, Daniel Bates, Chris Doran}\\
        \fontsize{11}{14}\selectfont
        Monumo Ltd, Cambridge, UK \\
        Email: \{jaroslaw.rzepecki, daniel.bates, chris.doran\}@monumo.com
        }

\begin{document}

\twocolumn[
  \begin{@twocolumnfalse}
    \maketitle

\noindent \textbf{Abstract.} We present a novel method for using Neural Networks (NNs) for finding solutions to a class of Partial Differential Equations (PDEs). Our method builds on recent advances in Neural Radiance Field research (NeRFs) and allows for a NN to converge to a PDE solution orders of magnitude faster than classic Physically Informed Neural Network (PINNs) approaches. \\

\noindent \textbf{Keywords:}  NeRF, PINNs, PDEs, Neural Networks, Electromagnetism, Maxwell Equations

  \end{@twocolumnfalse}
  \vspace{1.5em}
]


\section{Introduction}
\label{sec:introduction}
Partial differential equations (PDEs) are found throughout physics and engineering, but solving them is notoriously difficult. Numerical approaches such as Finite Element Methods (FEMs) simplify the problem by breaking it down into smaller pieces (``elements''), and making assumptions about the relationships within and between those pieces. This is effective for small problems, but does not scale well as the number of elements grows.

Recently, Physically Informed Neural Networks (PINNs) have exploited neural networks' ability to approximate functions to learn solutions to PDEs. They are trained by measuring how far the neural network's function is from a solution to the PDE, and computing a loss term accordingly. This can naturally be extended to solve multiple systems of PDEs simultaneously, but training a neural network can take a long time. Neural networks can also lead to a more compact and elegant functional representation of the solution to a PDE, instead of the brute force enumeration of positions and values typically used by FEMs.

Our method provides a faster alternative to PINNs by building on recent developments in Neural Radiance Fields (NeRFs). In particular, PDEs often tell us requirements of any path or surface in the domain, which allows us to efficiently generate arbitrary amounts of training data by simply generating random paths and/or surfaces and computing their required properties. We have also found that sampling the neural network's output along entire paths instead of individual points improves numerical stability and makes it easier for the network to converge to a result.



\section{Physically Informed Neural Networks}
\label{sec:pinns}
Neural Network based PDE solvers [Raissi et al.~\cite{raissi2019physics}, Lagaris et al.~\cite{lagaris1998artificial}, Sirignano et al.~\cite{sirignano2018dgm}, Hennigh et al.~\cite{hennigh2021nvidia}], after successful training, allow for nearly instantaneous solving of parametrized models/PDEs, which can benefit both manual (human in the loop) and fully automated optimization schemes. Additionally, at an expense of a more-complicated loss function, they make combination of multiple physical phenomena (i.e. different types of PDEs) relatively straightforward.

Neural Network solvers are based on two main observations: (1) Neural Networks are very good general function approximators and (2) it is difficult to find a solution to a PDE, however it is easy to check if a given candidate function does solve a PDE.

Following the notation from Hennigh et al.~\cite{hennigh2021nvidia} let us consider a general form of a PDE:
\begin{equation} \label{eq:pde_general_1}
\begin{array}{ll}
\mathcal{N}_{i}[u](\mathbf{x})=f_{i}(\mathbf{x}), & \forall i \in\left\{1, \cdots, N_{\mathcal{N}}\right\}, \mathbf{x} \in \mathcal{D} \\
\mathcal{C}_{j}[u](\mathbf{x})=g_{j}(\mathbf{x}), & \forall j \in\left\{1, \cdots, N_{\mathcal{C}}\right\}, \mathbf{x} \in \partial \mathcal{D}
\end{array}
\end{equation}
In the above equation $\mathcal{N}_{i}$s are differential operators, $\mathbf{x}$ is the set of independent variables defined over a bounded continuous domain $\mathcal{D} \subseteq \mathbb{R}^{D}, D \in\{1,2,3, \cdots\}$, and $u(\mathbf{x})$ is the solution to the PDE. $\mathcal{C}_{j}$s denote the constraint operators that cover the boundary and initial conditions. $\partial \mathcal{D}$ also denotes a subset of the domain boundary that is required for defining those constraints. We seek to approximate the solution $u(\mathbf{x})$ by a neural network $u_{n e t}(\mathbf{x})$.

Depending on the actual problem at hand and the type of the PDE, different neural network architectures can be used. Generally, a standard multi-layer fully-connected architecture that transforms the input $\mathbf{x}$ into $u_{n e t}(\mathbf{x})$ has been found to work well for a large class of problems.

The loss function needed to train the neural network can be constructed based on how well the neural network approximated solution $u_{n e t}(\mathbf{x})$ fulfils the PDE and its boundary conditions. To construct the loss function, first we define the PDE ($r_{\mathcal{N}}^{(i)}$) and constrains ($r_{\mathcal{C}}^{(j)}$) residuals:
\begin{equation} \label{eq:pde_nn_residuals}
\begin{aligned}
r_{\mathcal{N}}^{(i)}\left(\mathbf{x} ; u_{n e t}(\theta)\right) &=\mathcal{N}_{i}\left[u_{n e t}(\theta)\right](\mathbf{x})-f_{i}(\mathbf{x}), \\
r_{\mathcal{C}}^{(j)}\left(\mathbf{x} ; u_{n e t}(\theta)\right) &=\mathcal{C}_{j}\left[u_{n e t}(\theta)\right](\mathbf{x})-g_{j}(\mathbf{x}), 
\end{aligned}
\end{equation}
where $\theta$ is the set of neural network trainable parameters (weights and biases).

The final loss function, $\mathcal{L}_{\text {res }}(\theta)$, can now be defined as a weighted sum of all the residuals:
\begin{equation} \label{eq:pde_nn_loss}
\begin{aligned}
\mathcal{L}_{\text{res }}(\theta)={} & \sum_{i=1}^{N_{\mathcal{N}}} \int_{\mathcal{D}} \lambda_{\mathcal{N}}^{(i)}(\mathbf{x})\left\|r_{\mathcal{N}}^{(i)}\left(\mathbf{x} ; u_{n e t}(\theta)\right)\right\| d \mathbf{x}\\
+&\sum_{j=1}^{N_{\mathcal{C}}} \int_{\partial \mathcal{D}} \lambda_{\mathcal{C}}^{(j)}(\mathbf{x})\left\|r_{\mathcal{C}}^{(j)}\left(\mathbf{x} ; u_{n e t}(\theta)\right)\right\| d \mathbf{x}
\end{aligned}
\end{equation}

In practice, the integrals in Eq.~(\ref{eq:pde_nn_loss}) can be approximated by discrete sampling of the input space. 

A neural network trained using the Loss from Eq.~(\ref{eq:pde_nn_loss}) will theoretically converge to a function that is a solution to the given set of PDEs and their boundary conditions. In practice it has been shown (Hennigh et al.~\cite{hennigh2021nvidia}) to work well for many different problems including heat transfer, mechanical stress, diffusion and others. However, for this method to work one needs access to the gradients of the test function (which are usually given by a NN training framework) and it requires careful manual tuning of the weight functions ($\lambda_{\mathcal{N}}^{(i)}$, $\lambda_{\mathcal{C}}^{(j)}$) that control the relative significance of the loss components (i.e. PDE solution vs boundary condition). 

In particular for electromagnetic problems involving Neumann boundary conditions between materials with different magnetic (relative) permeability ($\mu_{air} \approx 1.0$, $\mu_{iron} \approx 5000$) we found that the boundary condition part of the loss function was not numerically stable (due to the orders of magnitude jump in function gradient requirement on the boundary) and it was difficult to choose its weighting in such a way that the network would converge to the full PDE solution.

\section{Neural Radiance Fields}
\label{sec:nerfs}
Neural Radiance Field methods provide a neural-network-backed method for synthesising new views of complex 3D scenes by optimizing an underlying volumetric scene function using a sparse set of input
views/images [Mildenhall et al.~\cite{nerf:mildenhall}]. The 3D scene is represented using a neural network (usually a fully-connected deep network). The input to the network is a 5D coordinate -- spatial location ($x, y, z$) and viewing direction ($\theta, \phi$). The output of the network is the view-dependent emitted radiance and the volume density at the corresponding input location. A 2D view (image) can be synthesised by querying 5D coordinates along camera rays and then using classic volume rendering techniques to project the output colors and densities into an image. This process involves integrating the radiance and density outputs along each camera ray to produce a final pixel in the 2D image.

The training data set contains a set of 2D images of the 3D scene we want to recreate, each image is taken with a known camera angle. For a given pixel for the training loss is a difference between the value of that pixel in the training image, and the colour of that pixel predicted by the NeRF model. To obtain the pixel colour $\boldsymbol{C}$ from the NeRF model we need to integrate the outputs (radiance $\boldsymbol{c}$ and density $\sigma$) of the neural network along the ray $\boldsymbol{r}(t)$ [Mildenhall et al.~\cite{nerf:mildenhall}]:

\begin{equation} \label{eq:radinace_integral}
C(\mathbf{r})=\int_{t_{near}}^{t_{far}} T(t) \sigma(\mathbf{r}(t)) \mathbf{c}(\mathbf{r}(t), \mathbf{d}) d t, 
\end{equation}
where:
\begin{equation} \label{eq:radinace_transmitt}
T(t)=\exp \left(-\int_{t_{near}}^{t} \sigma(\mathbf{r}(s)) d s\right)
\end{equation}
is the accumulated transmittance along the integrated ray up until point $t$. The integrals in Equations (\ref{eq:radinace_integral}, \ref{eq:radinace_transmitt}) are taken along a line (ray) parameterised with $t$, starting at $t_{near}$ and terminating at $t_{far}$:
\begin{equation} \label{eq:radiance_ray}
\mathbf{r}(t) = \mathbf{o} + t\mathbf{d}
\end{equation}
where, $\mathbf{o}$ is the ray origin (determined by the position of the pixel in a given 2D image/view) and $\mathbf{d}$ is the ray direction which is determined by the camera angle used when the given image/view was generated. In practice, both integrals are approximated by sampling and summation.

Total training loss can then be constructed as an average error of colour predictions of all the training pixels.

Note that in order for the neural network to correctly recreate the radiance and the density of the 3D scene it must be capable of capturing very rapid changes in its output in relation to its inputs. In particular the density output $\sigma$ will have its values change dramatically on the boundaries of solid objects in the scene. If the network was not capable of recreating those rapid density changes, than all the images/views recreated by the NeRF model would be blurry without proper definition of solid object boundaries.

To sum up -- a NeRF model is trained on a loss function that is based on a line integral(s) of the network outputs Eq.~(\ref{eq:radinace_integral}) and it needs to be able to properly capture rapid changes in the output values with respect to the inputs.

\section{Input encoding}
\label{sec:input_enc}
Transformation of the inputs to the neural network (input encoding) has been noted to play a crucial role for speed and accuracy of training both in the case of PINNs and NeRF. Hennigh et al.~\cite{hennigh2021nvidia} implement both Fourier based encoding and SiReN encoding introduced by Sitzmann et al.~\cite{encoding:siren}. Both of these encodings aim at increasing the ability of the network to learn important high frequency features. 

The work of M{\"{u}}ller et al.~\cite{encoding:instant_ngp} takes this a step further and focuses on adaptive encoding. This adaptive, multi-resolution hash-based encoding lead to several orders of magnitude speed-up in training of a set of graphics related tasks, including a NeRF model.

\section{Fast Integral PINN Solving}
\label{sec:intpinn}
In this section we describe our method for fast solving of a class of PDEs. For clarity of argument we will focus on solving magneto-static Maxwell equations, but the methodology can be easily applied to any type of PDE that can be represented in integral form.

A full set of general Maxwell equations:
\begin{equation} \label{eq:mawell_diff}
\begin{aligned}
&\nabla \times \mathbf{H}=\mathbf{J}_{c}+\frac{\partial \mathbf{D}}{\partial t} \\
&\nabla \times \mathbf{E}=-\frac{\partial \mathbf{B}}{\partial t} \\
&\nabla \cdot \mathbf{D}=\rho \\
&\nabla \cdot \mathbf{B}=0\\
\end{aligned}
\end{equation}
can be rewritten in integral form:
\begin{equation} \label{eq:mawell_int}
\begin{aligned}
&\oint \mathbf{H} \cdot d \mathbf{l}=\int_{S}\left(\mathbf{J}_{\mathrm{c}}+\frac{\partial \mathbf{D}}{\partial t}\right) \cdot d \mathbf{S} \\
&\oint \mathbf{E} \cdot d \mathbf{l}=\int_{S}\left(-\frac{\partial \mathbf{B}}{\partial t}\right) \cdot d \mathbf{S} \\
&\oint_{S} \mathbf{D} \cdot d \mathbf{S}=\int_{v} \rho d v \\
&\oint_{S} \mathbf{B} \cdot d \mathbf{S}=0 
\end{aligned}
\end{equation}
In the above equations, $\mathbf{H}$ is magnetic field strength, $\mathbf{B}$ is magnetic flux density, $\mathbf{E}$ is electric field strength, $\mathbf{D}$ is electric displacement field, $\rho$ is electric charge density and $\mathbf{J_{c}}$ is electric current density. The line integrals are taken along any closed loop, and the surface integrals are taken over any closed surface S.

For the magneto-static case, the integral equations reduce to:
\begin{equation} \label{eq:mawell_int_static}
\begin{aligned}
&\oint \mathbf{H} \cdot d \mathbf{l}=\int_{S}\mathbf{J}_{\mathrm{c}} \cdot d \mathbf{S} \\
&\oint_{S} \mathbf{B} \cdot d \mathbf{S}=0 
\end{aligned}
\end{equation}
with the relation between $\mathbf{H}$ and $\mathbf{B}$ given by permeability $\mu$ of a material at the given point is space: $\mathbf{B}=\mu\mathbf{H}$. In general, the value of $\mu$ for a given material is a function of $\mathbf{H}$ (material $\mathbf{B/H}$ curve).

In a 2D case that we consider here, we can further rewrite Eq.~(\ref{eq:mawell_int_static}) as:
\begin{equation} \label{eq:mawell_int_static_2d}
\begin{aligned}
&\oint_{L} \mathbf{H_{||}} \cdot d \mathbf{l}=\mathbf{J}_{enc} \\
&\oint_{L} \mathbf{B_{\perp}} \cdot d \mathbf{n}=0 
\end{aligned}
\end{equation}
where, $\mathbf{J}_{enc}$ is the total current enclosed by the close loop L, $d \mathbf{l}$ is the loop integration element parallel to the integration line and $d \mathbf{n}$ is an element perpendicular to $d \mathbf{l}$.

The integrals in Eq.~(\ref{eq:mawell_int_static_2d}) have to be fulfilled for any closed loop. A single loop integral is not enough to uniquely determine the $\mathbf{B}$ field, however if enough random loops are considered in the given domain, than it will be enough to find the solution. This makes it similar to the NeRF problem, where a single ray traced through the scene is not enough to determine the 3D radiance filed, but many rays combined are.

This conceptual similarity between NeRF ray integration and the integral form of Maxwell magneto-static equations is the basis of our method, which allows us to quickly train a neural network to find a solution to any PDE as long as it can be expressed in integral form.

\subsection{Method}
\label{sec:method}
Our goal is to find the $\mathbf{B}$ field for any $(x,y)$ point in the domain of interest, assuming that we know the material properties $\mu(x,y)$ and the current distribution ($\mathbf{J_{c}}$). To do so, we introduce a neural network $\mathcal{F}$ that maps input coordinates $(x,y)$ to $\mathbf{B}_{net}$. The architecture of the network and an additional encoding of the inputs can be chosen to best suit the problem at hand. We chose to use a fully connected architecture, with multi-resolution hash input encoding from M{\"{u}}ller et al.~\cite{encoding:instant_ngp}.

\begin{figure}[t]
    \centering
    \includegraphics[width=\linewidth]{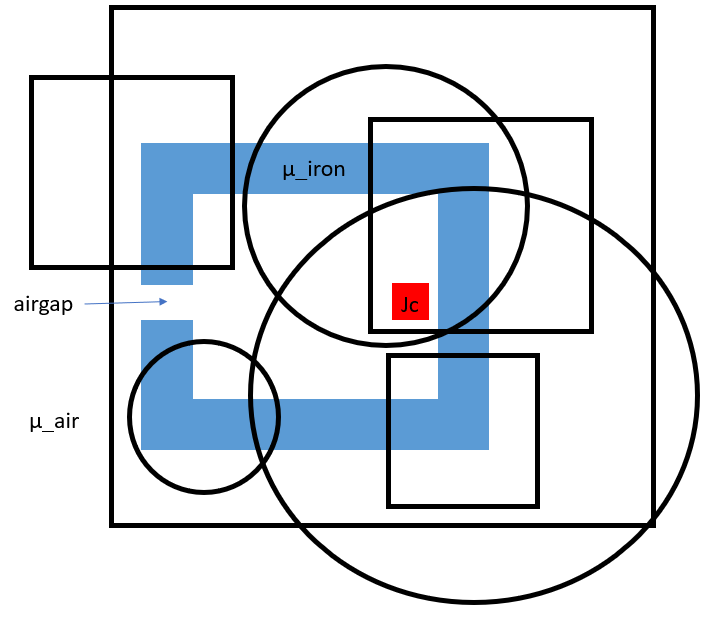}
    \caption{Test problem setup. Iron core electromagnet ($\mu\_iron$), surrounded by air. Force is provided by a wire (red square) with current density Jc. Black lines show example integration paths.}
    \label{fig:paths}
\end{figure}

The training loss function is constructed by calculating the integrals from Equations~(\ref{eq:mawell_int_static_2d}) along many randomly chosen closed paths within the spatial domain of interest. For implementation simplicity we have used integration paths that are either squares or circles of random dimensions, see Fig~\ref{fig:paths}. For a single path L, the loss function is:
\begin{equation} \label{eq:maxwell_loss}
\mathcal{L}_{\text {L }}(\theta)=\left\|\oint_{L} \mathbf{B^{\perp}_{net}} \cdot d \mathbf{n}\right\| +
\left\|\oint_{L}\frac{\mathbf{B^{||}_{net}}}{\mu} \cdot d \mathbf{l}-\mathbf{J}_{enc}\right\|
\end{equation}
where $\theta$ are trainable parameters of the neural network and a training batch consists of multiple integration paths.

By training the network with the above loss function on enough different integration paths, the network will learn the solution $\mathbf{B}_{net}$ to Eq.~(\ref{eq:mawell_int_static_2d}) for any point within the domain.

\section{Results}
\label{sec:results}
We have tested our method on a simple magneto-static setup of a ``horse-shoe'' electromagnet. This setup consist of a C-shaped iron core (light blue, $\mu\_iron$ in Fig~\ref{fig:paths}), surrounded by air ($\mu\_air$). The force component is provided by a wire with current density Jc (red square in Fig~\ref{fig:paths}).

To model the variable $\mu\_iron$, we use a simple piecewise-linear approximation to the true $\mathbf{B/H}$ curve. We use two values of $\mu\_iron$, one when the iron is saturated with magnetic flux and one when it is not saturated, and we determine the saturation level by sampling the neural network to estimate $\mathbf{B}$.

\begin{figure}[t]
    \centering
    \begin{subfigure}[t]{0.4\linewidth}
    \includegraphics[width=\linewidth]{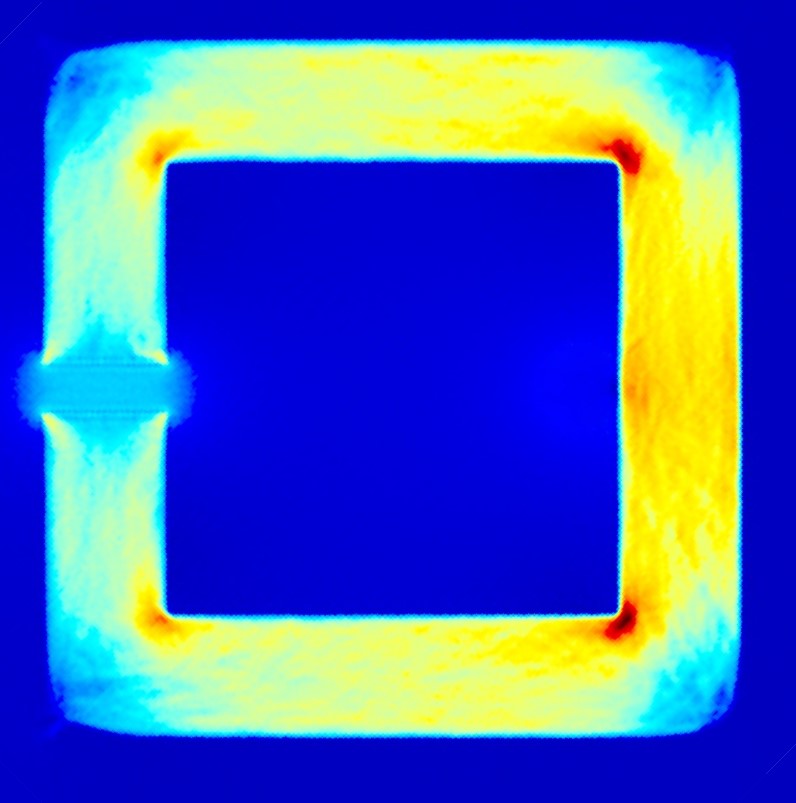}
    \caption{Magnetic flux density obtained with our method.}\label{fig:horseshoe_ours}
    \end{subfigure}
    \quad
    \begin{subfigure}[t]{0.4\linewidth}
    \includegraphics[width=\linewidth]{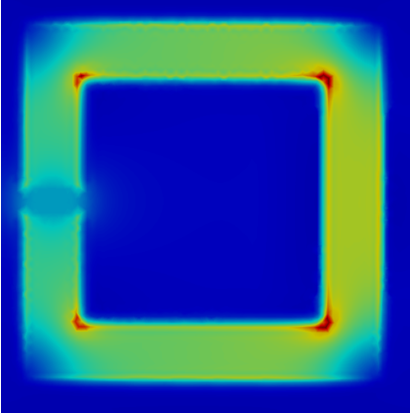}
    \caption{Reference magnetic flux density obtained by FEM solution (Elmer).}\label{fig:horseshoe_reference}
    \end{subfigure}
    \caption{Comparison of magnetic flux magnitude of our method against reference solution from FEM solver (Elmer). Blue corresponds to no flux, red is maximum flux (intermediate colour mappings have not been aligned).} 
    \label{fig:compare}
\end{figure}

\begin{figure}[t]
    \centering
    \includegraphics[width=\linewidth]{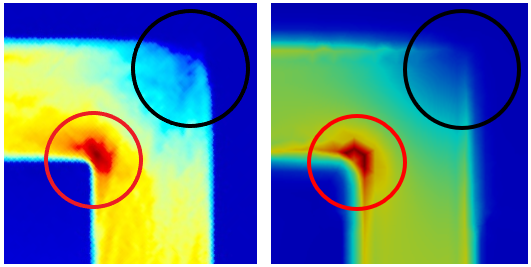}
    \caption{Corner zoomed comparison between our solution (left) and reference (right). Both solutions show high flux density on the inner corner (red circle), and low flux at the outer corner (black circle).}
    \label{fig:corner_comp}
\end{figure}

\begin{figure}[t]
    \centering
    \includegraphics[width=\linewidth]{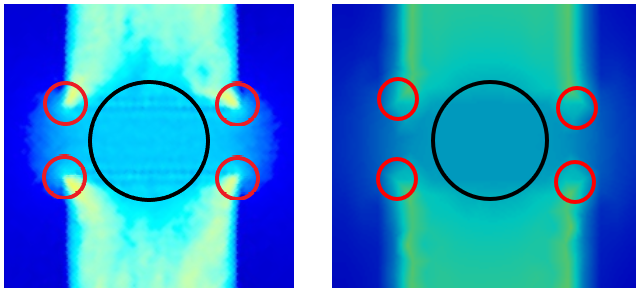}
    \caption{Airgap zoomed comparison between our solution (left) and reference (right). Both solutions show high flux density on the airgap iron corners (red circles), and lower flux in the middle of the gap (black circle).}
    \label{fig:airgap_comp}
\end{figure}

Figure~\ref{fig:compare} shows the magnitude of magnetic flux B obtained by our method compared to reference solution from the Elmer FEM solver (https://github.com/ElmerCSC/elmerfem). Please note that the electric current driving both cases is slightly different -- for FEM there is a current density over a coil surface, for our solution we have used a single wire. The singularity associated with a point wire has been removed from Fig~\ref{fig:horseshoe_ours}.

Thanks to the same input encoding that sped up NeRF training by orders of magnitude (M{\"{u}}ller et al.~\cite{encoding:instant_ngp}) the solution shown in Fig~\ref{fig:horseshoe_ours} has been obtained after only about 30 seconds of training on a high-end laptop. This is considerably faster than existing PINN methods, but slower than standard FEM approaches. However, we believe that neural approaches will ultimately be more flexible as they can more naturally be extended to accept design space parameters as inputs, and once trained, will be able to instantly generate solutions for new designs.

Our solution matches the reference solution very closely, including key critical properties of the magnetic flux density in this setting. Namely: 
\begin{enumerate}
\item The flux ``cuts corners'' -- the flux density in high close to the inner corners of the iron core and very low at the outer corners -- see Fig~\ref{fig:corner_comp}.
\item The flux distribution in the airgap has the characteristic v-shape, with higher density at the corners of the iron -- see Fig~\ref{fig:airgap_comp}.
\item Generally all the flux is contained within the iron.
\end{enumerate}

\section{Conclusions}
\label{sec:conclusions}
By transforming the differential form of Maxwell equations for magneto-statics (Eq.~\ref{eq:mawell_diff}) to a closed loop integral form (Eq.~\ref{eq:mawell_int_static_2d}), and by applying the input encoding from the latest NeRF break-through paper (M{\"{u}}ller et al.~\cite{encoding:instant_ngp}) we managed to train a neural network to solve the magneto-static problem  orders of magnitude faster than if using classical PINN approach (i.e.\ training with differential form loss function~\ref{eq:pde_nn_loss}). Our method can be used each time a PDE (or other type of equation) can be transformed to a line integral (closed or not). An extension to 3D would require changing the integration lines into surfaces but the rest of the methodology would remain the same. The integral form of the equations helps deal with numerical instabilities and at the same time allows us to take advantage of novel input space encoding introduced in NeRF research. 

\subsection{Future research}
\label{sec:future}
An interesting next step would also be to try to train the network using parametrization of the input geometries (i.e. iron shapes) and current densities. This would lead to a PDE solution that can generalize across a set of geometries/currents and would therefore allow for usage in optimization routines as a extremely fast approximation to FEM solution.

\section*{Acknowledgement}
\label{acknowledgement}
This work has been carried out as part of Monumo Ltd.\ research and development efforts.\\

\par
\printbibliography

\end{document}